\documentclass[11pt]{article}

\usepackage{eacl2017}
\usepackage{times}
\usepackage{url}
\usepackage{latexsym}

\usepackage{amsmath,amssymb}
\usepackage{graphicx}
\usepackage[capitalize]{cleveref}
\usepackage{hyphenat}
\usepackage{booktabs}
\usepackage{tabularx}
\usepackage[font=small]{caption}
\usepackage{multirow}

\eaclfinalcopy % Uncomment this line for the final submission
\def\mdim{multi\hyp dimensional}
\def\Mdim{Multi\hyp dimensional}

\title{Towards Learning Transferable Conversational Skills using Multi-dimensional Dialogue Modelling}
\author{Simon Keizer and Verena Rieser \\
  Interaction Lab \\
  Heriot-Watt University, Edinburgh (UK) \\
  {\tt \{s.keizer, v.t.rieser\}@hw.ac.uk} \\}
\date{}

\begin{document}

\maketitle

\begin{abstract}
Recent statistical approaches have improved the robustness and scalability of spoken dialogue systems.  However, despite recent progress in domain adaptation, their reliance on in-domain data still limits their cross-domain scalability.  In this paper, we argue that this problem can be addressed by extending current models to reflect and exploit the multi-dimensional nature of human dialogue.  We present our multi-dimensional, statistical dialogue management framework, in which transferable conversational skills can be learnt by separating out domain-independent dimensions of communication and using multi-agent reinforcement learning.  Our initial experiments with a simulated user show that we can speed up the learning process by transferring learnt policies.
\end{abstract}

\section{Introduction}\label{sec:intro}

Virtual personal assistants, such as Siri, Cortana, Google Now, and Alexa, have made commercial use of interactive spoken language technology.  However, commercial exploitation of advanced spoken dialogue technology requires new methods for cost-effective development and efficient adaptation to new domains.  In this paper, we argue that this problem can be addressed by taking a \emph{\mdim\ }approach.

Current systems focus almost exclusively on the primary task underlying the conversation, for example travel booking or seeking tourist information.  The behaviour resulting from such an approach is quite different from natural human dialogue, where several other aspects besides the task itself are addressed as well, such as giving and eliciting feedback, following social conventions, and managing turn-taking and timing.  Humans frequently perform \emph{multi-functional} utterances, where several of these aspects, or \emph{dimensions}, are addressed simultaneously \cite{Bunt:2011et}.  Consider the following example interaction (annotated with different functions for each turn):
\begin{center}
\small
\begin{tabular}{@{}l@{\;}l@{}}
\toprule
\textbf{Usr}: & \textsl{\textbf{Hello, I am looking for a \underline{cheap} \underline{Indian} restaurant}} \\[1mm]
\multicolumn{2}{l}{\hspace{3mm}\textsc{Social:Greet; Task:Inform; Turn:Release}} \\[2mm]
\midrule
\textbf{Sys}: & \textsl{\textbf{Okay, let me see, \dots}} \\
\multicolumn{2}{l}{\hspace{3mm}\textsc{AutoPositive; Time:Pausing; Turn:Keep}} \\[2mm]
\textbf{Sys}: & \textsl{\textbf{The Rice Boat is an \underline{Indian} restaurant}} \\[1mm]
                 & \hspace{3mm}\textsl{\textbf{in the \underline{cheap} pricerange}} \\[2mm]
\multicolumn{2}{l}{\hspace{3mm}\textsc{Auto-feedback:Inform; Task:Inform}} \\
\bottomrule
\end{tabular}
\end{center}
The user both greets the system and tells the system they want a cheap Indian restaurant, before releasing the turn; the system then takes the turn with positive feedback and indicates that it needs more time to retrieve the requested information; in the second part the system both provides this information and gives feedback about understanding the user's question (underlined).

Following the notion of multi-dimensionality of dialogue as described by Bunt~\shortcite{Bunt:2011et} and early exploratory work on \mdim\ dialogue management by Keizer and Bunt~\shortcite{Keizer:2006vn,Keizer:2007ve}, we present a new framework for statistical dialogue management which explicitly accounts for these different dimensions of communication.  By separating out domain-independent dimensions, our approach has the potential to learn a set of transferable conversational skills, enabling more efficient cross-domain adaptation.  

In \cref{sec:md-dial} we discuss the theoretical background of our approach, followed in \cref{sec:md-pomdp-dial} by its embedding into a statistical dialogue system framework.  In \cref{sec:sds-impl} we present the first implementation of our \mdim\ statistical dialogue manager, including components for state monitoring and action selection, and the user simulator used for testing, training and evaluation.  We then present preliminary experiments in \cref{sec:sim-exp}, demonstrating the potential of our method for cross-domain transfer.  We conclude the paper in \cref{sec:concl}.

\section{\Mdim\ Dialogue Modelling}\label{sec:md-dial}

In Bunt's account of multi-dimensionality in dialogue, utterances are represented as combinations of dialogue acts from a \mdim\ dialogue act taxonomy, thus accounting for their multifunctional nature \cite{Bunt:2011et}.  This taxonomy, which is part of the ISO standard for dialogue act annotation \cite{ISO-SemAnnot}, includes the following 9 core dimensions: \textsl{Task/Activity}, \textsl{Auto-}, and \textsl{AlloFeedback}, \textsl{Turn-}, and \textsl{TimeManagement}, \textsl{Partner-} and \textsl{Own Processing Management}, \textsl{Discourse Structuring}, and \textsl{Social Obligations Management}.  In producing utterances, dialogue partners select one or more dialogue acts, at most one from each dimension.  The second system utterance in the example of \cref{sec:intro} is the result of the system selecting an answer act in the Task dimension and an inform act in the AutoFeedback dimension, which are then combined and realised as a single multi-functional utterance.  However, some combinations of dialogue acts can only be realised sequentially in a natural language, such as the greeting and the question in the user utterance of the example in \cref{sec:intro}, even though these acts were selected simultaneously by the agent.

A key feature of the dialogue act taxonomy we aim to exploit, is that all dimensions except Task are domain-independent.  A dialogue agent uses the same dialogue acts for managing the turn-taking process, or for following social conventions such as greeting and thanking, regardless of the underlying task or activity.  Moreover, we believe that to some extent, the strategies for selecting these domain-independent dialogue acts can be largely transferred across tasks/activities.  When changing from one task to the other, a dialogue participant does not need to learn from scratch how to achieve mutual understanding through feedback dialogue acts; they merely need to adapt their strategy to the new circumstances.  Turn management for example, will depend on the communicative settings of the dialogue, i.e., whether the dialogue is a telephone conversation (speech only) or face-to-face (speech and gestures).  In safety critical domains, giving and eliciting feedback will be more explicit.  In more informal settings, or domains where for example empathy is important, the strategy for handling social conventions will be more elaborate, or at least different.  

Our dialogue system follows the ISO standard in the same way as Keizer and Bunt~\shortcite{Keizer:2006vn,Keizer:2007ve}, featuring multiple dialogue act agents, each dedicated to selecting dialogue acts from one of the dimensions, and a process of evaluating combinations of dialogue act candidates.  However, in our framework this is incorporated into a statistical dialogue manager, where the action selection policies are jointly optimised using multi-agent reinforcement learning.  This combination of multi-dimensional modelling and machine learning opens up the opportunity to use transfer learning methods for more efficient cross-domain adaptation of dialogue systems.

\section{\Mdim\ POMDP-based dialogue management}\label{sec:md-pomdp-dial}

\begin{figure}[htb]
\centering
\includegraphics[width=.95\columnwidth]{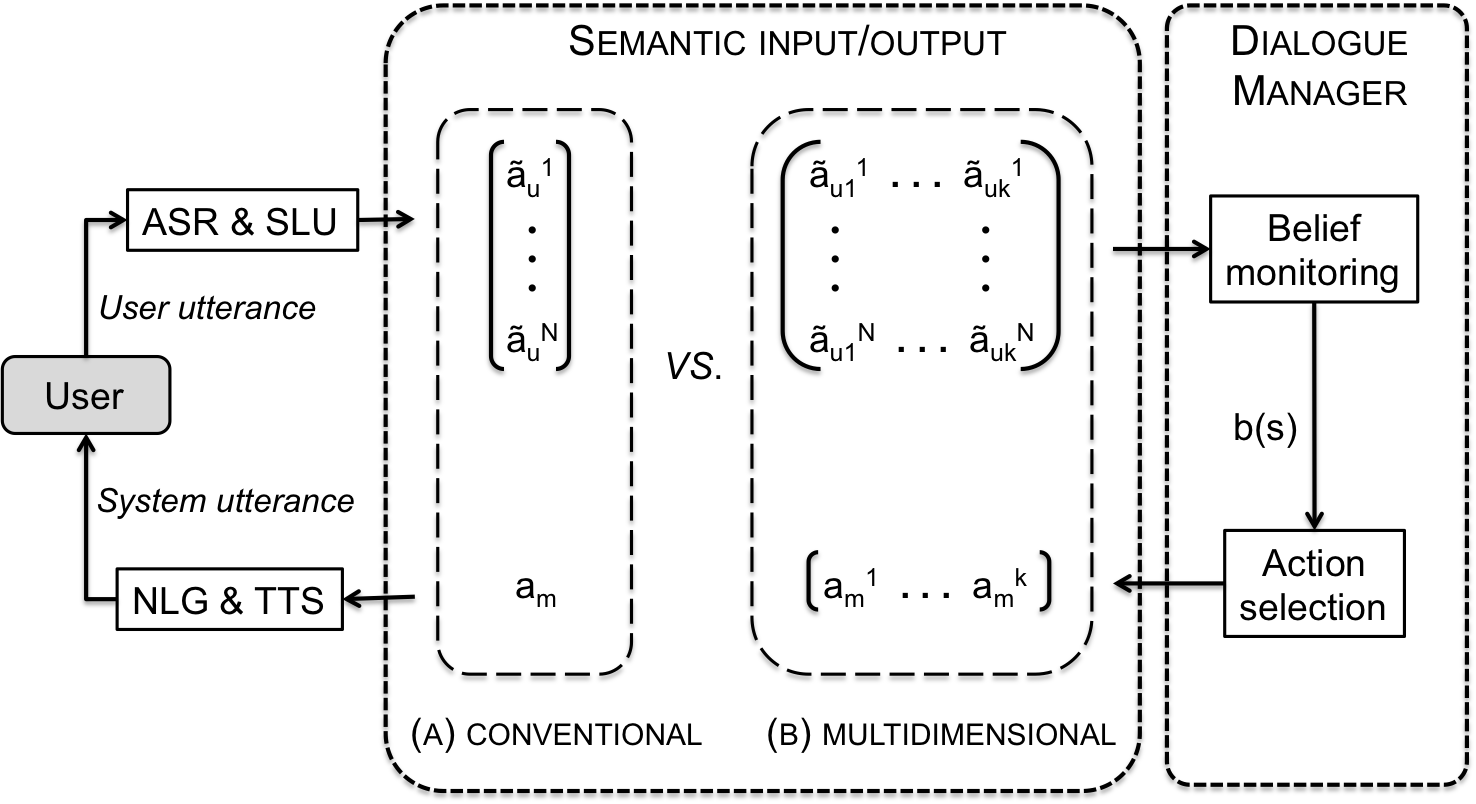}
\caption{Typical dialogue system architecture, contrasting a conventional statistical dialogue manager with a \mdim\ version.}
\label{fig:sys_diag}
\end{figure}

Recent advances in statistical dialogue systems have investigated Reinforcement Learning (RL) to optimise dialogue policies \cite{rl:springer11,young-etal-ieee-review}.  The underlying problem is modelled as a Partially Observable Markov Decision Process (POMDP) to account for uncertainty introduced by automatic speech recognition and spoken language understanding (ASR \& SLU).  A conventional POMDP-based spoken dialogue system typically consists of a pipeline of components for speech recognition and understanding (ASR \& SLU), dialogue management (DM), and natural language generation and speech synthesis (NLG \& TTS), see \cref{fig:sys_diag}, where the DM consists of \emph{belief monitoring} (updating the \emph{belief state} $b(s)$, i.e., a distribution over dialogue state hypotheses, based on an N-best list of user act hypotheses $\tilde{a}_u^i$), and \emph{action selection} (deciding which system act $a_m$ to generate, given the current belief state).  By combining probabilistic belief monitoring with reinforcement learning of dialogue policies, these systems have been demonstrated to be more robust to speech processing errors and more scalable to larger application domains. 

A major limitation of data-driven approaches to spoken dialogue systems is their reliance on substantial amounts of (annotated) data in the target domain.  As the number of application domains is growing every day, accelerated by the emergence of the Internet of Things (IoT) in particular, new methods for cost-effective development of conversational interfaces for these domains are needed.  More recently, researchers have started to address this issue by looking at \emph{transfer learning} techniques \cite{Taylor:2009ur,Pan:2010dm,Lazaric:2012}, with the aim to speed up learning dialogue models and policies for a target domain by leveraging data and/or knowledge from a source domain.  Recent domain adaptation work, however, has primarily focused on identifying and exploiting similarities between domain ontologies in slot-filling task domains.  Ga\v{s}i\'{c} et al \shortcite{gasic-EtAl:2013:SIGDIAL} used Gaussian Process Reinforcement Learning (GPRL) to adapt a dialogue policy to a new slot being added to the domain.  Since their approach relies on correlations between belief states rather than the belief states themselves, such adaptation is feasible, as long as the correlations are sufficiently similar between the domains.  Using the GPRL framework extended with a Bayesian committee machine, they have also demonstrated successful transfer in a multi-domain setting, where the domains have different, but overlapping sets of slots \cite{gasic-EtAl:2015:ASRU}.  In similar multi-domain settings, transfer learning methods have been developed for state tracking \cite{Mrksic_ea-2015} and natural language generation \cite{Wen_ea-2016}.

Rather than focusing on the domain ontology and the task, our proposed multi-dimensional framework distinguishes domain-independent dimensions such as social obligations management and time management, which can be transferred directly between domains.  These transferable skills are trained jointly in one domain, and can be re-used and adapted in a new domain.  Task/domain oriented approaches as used in the domain extension and multi-domain settings discussed above, might be used within our framework as well.  In that case, we not only transfer domain-independent policies, but also the domain-specific policy associated with the task dimension.

\begin{figure}[htb]
\centering
\includegraphics[width=.9\columnwidth]{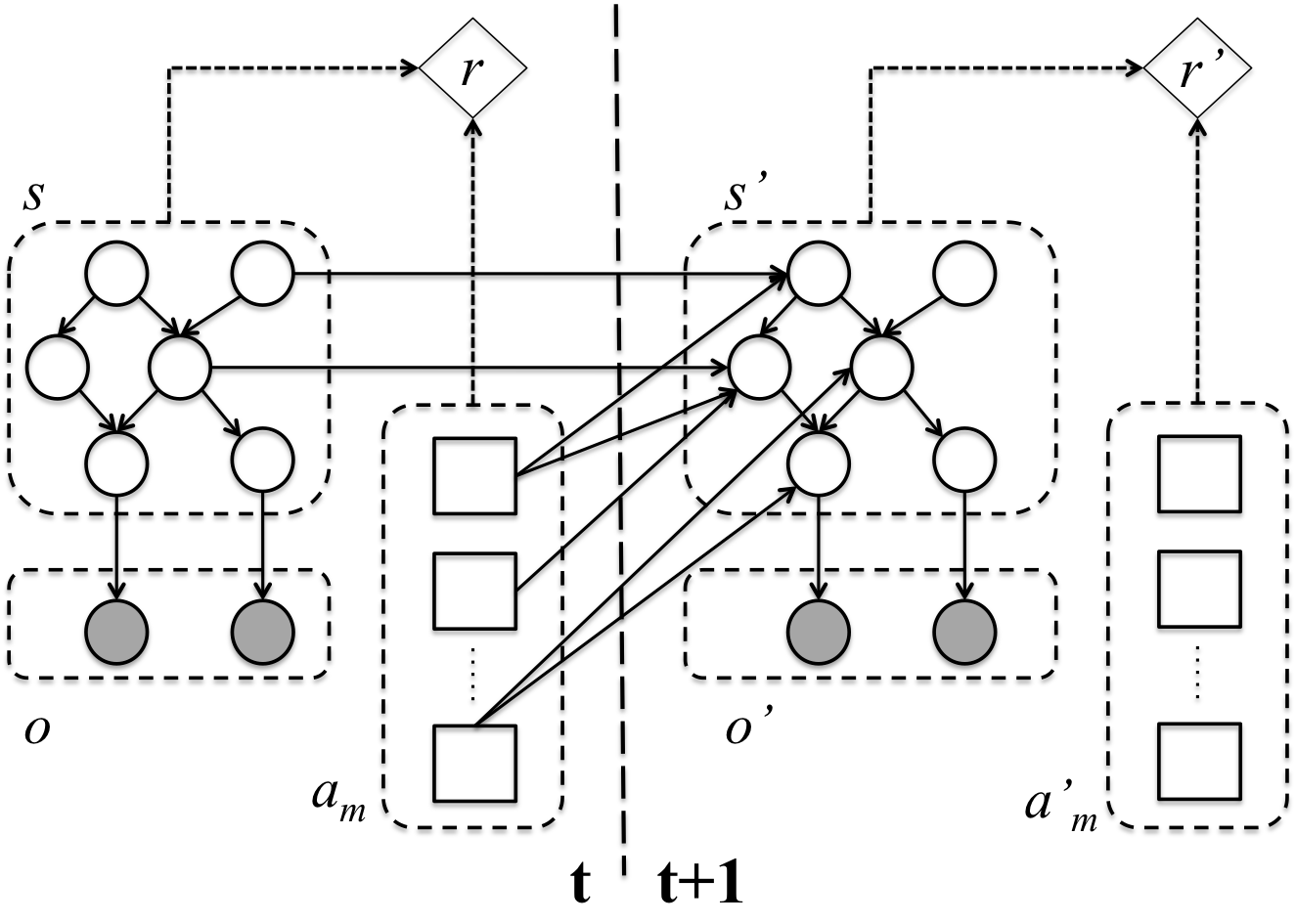}
\caption{Graphical model of a {POMDP} based SDS with factored action space.}
\label{fig:sds_pomdp_fa}
\end{figure}

Instead of selecting one dialogue act $a_m$ out of a single set of possible acts in each turn (the `conventional' setting A in \cref{fig:sys_diag}), our proposed DM selects responses that consist of combinations of dialogue acts $a_m^i$ (the `multidimensional' setting B in \cref{fig:sys_diag}).  The multi-dimensional POMDP model can also be represented with the graphical model shown in \cref{fig:sds_pomdp_fa}, which incorporates multiple action nodes, each associated with actions in one dimension, and affecting different sets of state variables.  A naive alternative to this factorisation of the action space would be to collapse the dimensions of dialogue acts back into a single set of actions, and then follow the conventional approach.  However, under this architecture the state-action space would grow exponentially and therefore unlikely to tractably accommodate the proposed richness of interaction.

\section{System implementation}\label{sec:sds-impl}

We have created a generic statistical dialogue manager for slot-filling domains, adopting many design features of the POMDP systems described in \cite{Young:2010vy,Thomson:2010dg}).  The dialogue manager consists of a probabilistic state monitoring model and an MDP-based action selection model.  To test, train and evaluate the dialogue manager, we have built an agenda-based user simulator based on \cite{Schatzmann:2007uc} and a basic error model, based on \cite{Thomson:2012wn}.  The user simulator generates dialogue acts in response to the dialogue manager, following a randomly selected user goal.  The error model then generates from this `true' user act an n-best list of user act hypotheses with confidence scores, to be passed to the dialogue manager.  Our simulated experiments have been carried out for the restaurant information domain, containing 4 `informable' slots ({\tt foodtype, pricerange, area, near}), 5 `requestable' slots ({\tt name, phonenumber, address, price, postcode}), and a database of 149 restaurants in Cambridge (UK).

\subsection{State Monitoring}

The dialogue state representation follows directly from the domain ontology and consists of user goal belief states for each of the informable slots (multinomial distributions over the slot values), beliefs about whether a requestable slot is indeed requested by the user (Bernoulli distributions), and other relevant information such as the dialogue history (previous dialogue acts), a list of database entities matching the user goal top hypothesis, and the database entity under discussion (if any).

The user goal beliefs $b(s,v)$ are updated as follows:
\begin{equation}\small
b'(s,v) = \left\{
  \begin{array}{ll}
  c(s,v)                    & \mbox{\small\; if evidence seen} \\
                               & \mbox{\;\; for the first time} \\[2mm]
  c(s,v) \cdot b(s,v) & \mbox{\; otherwise}
  \end{array}
  \right.
\end{equation}
where $(s,v)$ is a slot-value pair, $c(s,v)$ is a confidence score on evidence about a slot-value pair in the input n-best list of user act hypotheses.  This relatively simple belief tracker supports accumulation of evidence for slot values across multiple turns, where the slots are treated as independent.

Orthogonal to belief tracking, we also track grounding states (such as {\tt user\_informed} and {\tt system\_confirmed}) of user goal item hypotheses, which are updated according to a finite state machine similar to the model used in \cite{Young:2010vy}, based originally on \cite{Traum:Phd1994}.
% (\cite{Bunt_ea:sigdial07} describes a more elaborate grounding model for the ISO dialogue act taxonomy).

\subsection{Action Selection}

Based on the updated dialogue state, the dialogue manager selects response dialogue acts using one or more MDPs, each of which uses state features extracted from the full dialogue state and selects a summary action (e.g., `recommend a venue' or `ask slot preference') using a trainable policy, to be mapped back to a full dialogue act using information from the dialogue state (e.g., which venue to recommend or which slot to ask about).  The MDPs are trained using Monte Carlo Control reinforcement learning with linear value function approximation.  The reward signal is provided by the user simulator, assigning a score of -1 for each turn and a score of +30 when the user's goal is satisfied.

The MDPs consist of states $s \in S$, actions $a \in A$, and a policy $\pi:S \rightarrow A$ which maps states to actions.  The policy is based on the state action value function $Q:S \times A \rightarrow \mathbb{R}$ which approximates the long term cumulative reward when taking action $a$ in state $s$ and following the policy onwards.  During training we use $\epsilon$-greedy action selection.  The Q-values are approximated by a linear function of the state features $\phi_i(s)$:
\begin{equation}
Q(s,a) = \sum_i \theta_{i,a} \cdot \phi_i(s)
\end{equation}
After each dialogue/episode, the weight vectors $\mathbf{\theta}_a$ for each action $a$ are updated using gradient descent, minimising the squared difference between the current value estimates $Q(s,a)$ and the cumulative discounted rewards $R_t=\sum_{k=t}^{T-1} \gamma^{k-t} \cdot r_k$ for each visited state-action pair $(s_t,a_t)$ in the episode ($t = 0,\dots,T-1$), where $r_k$ are the immediate rewards received after each visited state action pair, and $\gamma=0.95$ is the discount factor.

\section{Preliminary experiments in simulation}\label{sec:sim-exp}

As a first proof-of-concept experiment, we have created a one-dimensional and a multi-dimensional version of our dialogue manager, which generate dialogue acts from the same action set.  Using the simulated user, we have carried out extensive policy optimisation experiments and compared the two systems.

\subsection{Experimental setup}

The two versions of the dialogue manager were created as follows.  The one-dimensional version uses a single MDP model, using an action set of 7 possible summary actions.  The multi-dimensional version uses three MDP models, corresponding to the dimensions \emph{Task} (5 actions, including asking for user preferences, making recommendations, presenting restaurant information), \emph{AutoFeedback} (3 actions, including asking clarification questions), and \emph{SocialOblMan} (2 actions, including goodbye acts).  The selected summary actions are combined into single system dialogue acts in a rule-based manner \cite{Keizer:2007ve}, ensuring the same range of output dialogue acts as the one-dimensional version.  For example, negative feedback acts cancel task acts, and goodbye acts are kept only if no candidate acts in the task dimension were selected (`null' actions).  In this restricted setting, the multi-dimensional version is expected to be more challenging to train, given the larger action space: $5\!\times\!3\!\times\!2=30$ action combinations versus $7$ actions.  \Cref{tab:mdp-action-stats} shows a description of the 7 actions in the one-dimensional system, the dimension of the resulting dialogue act, and the number of action combinations in the multi-dimensional system that map to this system act.  Since all dialogue act candidates are cancelled in the presence of negative feedback, all $5 \times 2 = 10$ combinations of the negative feedback act with Task and SocialOblMan acts are mapped to a negative feedback output act (see action index 0 in \cref{tab:mdp-action-stats}).  On the other hand, a returnGoodbye act is only allowed in combination with a `null' act from the task agent and if no negative feedback act is generated, leaving only 2 combinations (see action index 5 in \cref{tab:mdp-action-stats}).
\begin{table*}[tb]
\centering
\setlength{\tabcolsep}{3pt}
\begin{tabular}{ l c c }
\toprule
{\bf Action index \& description} & {\centering\bf  Dimension} & \parbox[c]{23mm}{\centering\bf \# Action combinations} \tabularnewline
\midrule
  0 -- negative feedback (``could you repeat that please?'')  &  AutoFeedback  & 10  \tabularnewline
  1 -- propositional question feedback (``did you say/mean \dots?'')  &  AutoFeedback  &  1  \tabularnewline
  2 -- answer to setQuestion w.r.t. task (``The address is \dots'') &  Task  &  4  \tabularnewline
  3 -- answer to propQuestion w.r.t. task (``Yes'', or ``No, it serves \dots'')  &  Task  &  4  \tabularnewline
  4 -- venue recommendation (``\dots is a nice place in the city centre'')  &  Task  &  4  \tabularnewline
  5 -- returnGoodbye act (``goodbye!'') --- closes the dialogue  &  SocialOblMan  &  2  \tabularnewline
  6 -- setQuestion w.r.t task (``What kind of food do you like?'')  &  Task  &  4  \tabularnewline
\bottomrule
\end{tabular}
\caption{Specification and quantitative comparison between one- and multi-dimensional MDP action sets.}\label{tab:mdp-action-stats}
\end{table*}

\subsection{Policy optimisation}

In all our policy optimisation experiments, 10 independent training runs have been carried out, and the evaluation results are averages over the 10 corresponding policy evaluations.  The one-dimensional system was trained over 40k dialogues with an exploration rate linearly decaying from $\epsilon=0.4$ to $\epsilon=0$ and a fixed learning rate of $\alpha=0.001$.  The multi-dimensional system was trained using the same settings, but now running the three MDP models simultaneously and updating their policies based on the same reward function.  This training process involves implicit coordination between the policies, within the restrictions of the combination rules.  For example, the task policy learns to stop making recommendations when the user is satisfied and says goodbye, whereas the social policy learns to respond to the user saying goodbye act and thus end the dialogue, but not before the task is completed.

\begin{figure}[htb]
\centering
\includegraphics[width=.45\textwidth]{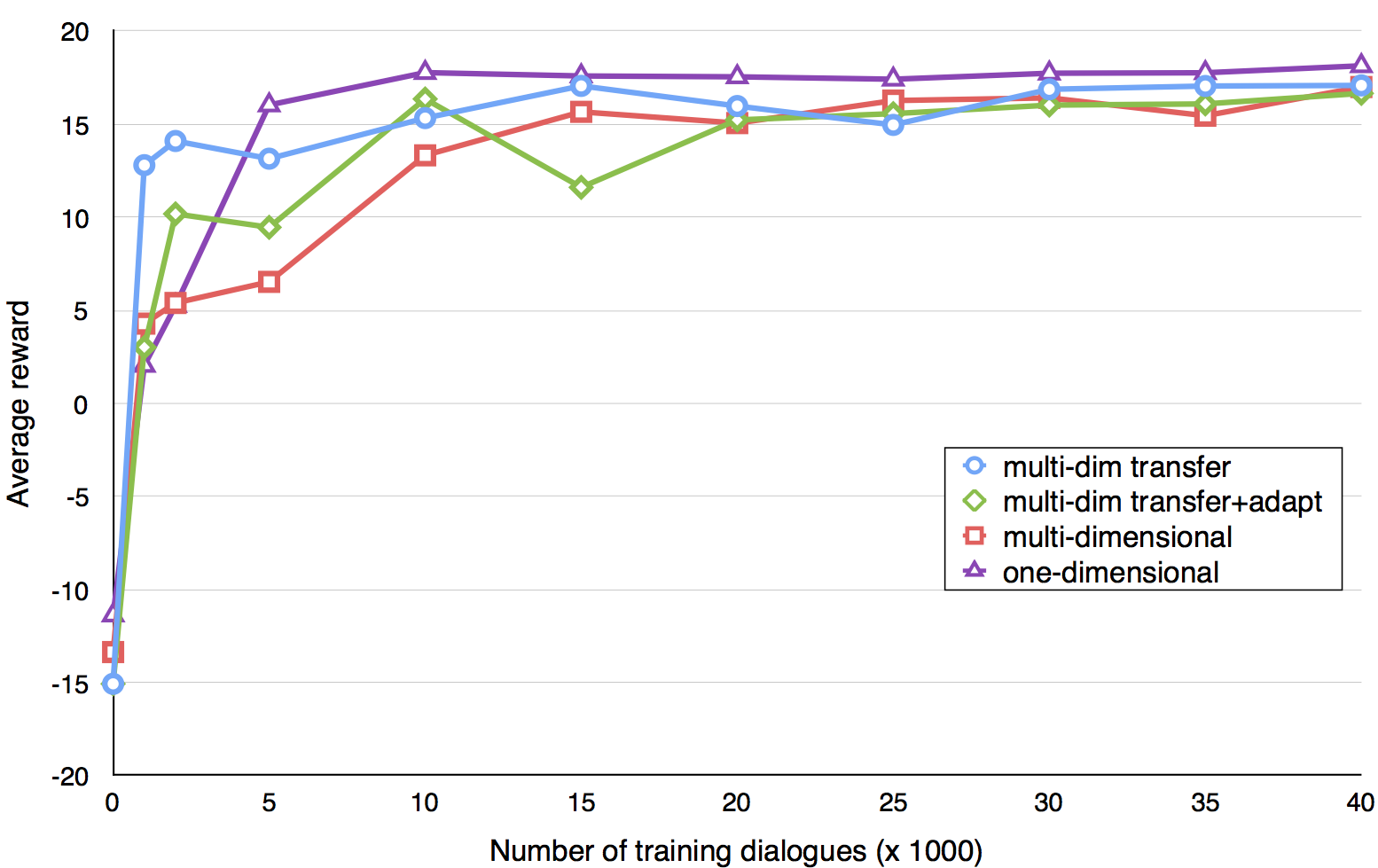}
\caption{Policy evaluation results of the one- and multi-dimensional systems in terms of average success rate at different training stages (20\% error rate was used throughout).}
\label{fig:curves}
\end{figure}

The learning curves in \cref{fig:curves} show the performance of trained policies at different training stages, where each data point represents the average reward over 3000 evaluation dialogues.  As expected, the one-dimensional system (purple, with triangular markers) achieves higher rewards than the multi-dimensional system (red, with square markers), in particular in the early stages of training.  However, after around 25k training dialogues, they have converged to similar performance levels (average reward 17--18; average dialogue length 11; average success rate 94--97\%).

After jointly optimising the three MDP policies, two domain-independent policies have been obtained that have the potential to be re-used in a new domain.  To demonstrate this potential in a first preliminary test without actually creating a new domain, we re-trained the dialogue manager in the same domain by retaining the trained auto-feedback and social obligations management policies (as if they were trained in a different source domain) and training the task policy from scratch (for the `new' target domain).  This domain transfer exercise was carried out in two settings: 1)~\textsl{multi-dim transfer}: only updating the task policy, i.e., keeping the trained domain-independent policies fixed, and 2)~\textsl{multi-dim transfer+adapt}: updating all three polices during training, i.e., adapting the trained domain-independent policies to the `new' domain.  The effectiveness of domain transfer is demonstrated by the corresponding learning curves in \cref{fig:curves}, which show improved performance levels at the earlier stages of training in comparison to the non-transferred multi-dimensional system.  Setting 1 (blue, with circular markers) shows clear and consistent improvement, whereas the improvement in setting 2 (green, with diamond markers) is more modest and training seems less stable (see the dips in performance at the 5k and 15k stages).  At the very early training stages, we even see improvements in comparison to the one-dimensional system.

\subsection{Discussion}\label{sec:disc}

Although the results of our initial experiments are encouraging, the next step of course is to extend our multi-dimensional system by refining the MDP models and allowing for system responses containing multiple dialogue acts.  For example, combinations of task and auto-feedback acts will be considered, as in the example dialogue exchange in \cref{sec:intro}, for which the auto-feedback MDP model will be extended to include decisions about which user-provided information the agent should give feedback about.  Our hypothesis is that a one-dimensional solution for generating such dialogue act combinations is less scalable.

Although some coordination between the dimensions will still be required, the MDP agents are also planned to be more independent.  To accommodate this, the reward function will also be decomposed into dimension-specific components.  For example, the SocialOblMan MDP agent can be extended with actions for apologies and responses to thanking acts, and trained based on a combination of an overall reward signal shared between all agents and a reward signal related specifically to social conventions and which the other agents will not receive or use.

In the above transfer experiment, adaptation of the domain-general policies was not necessary, since the target domain was identical.  For new domains, however, adaptation will be needed, for example safety critical domains where more explicit feedback is required, or informal domains where social interaction is more appropriate.

In the one-dimensional version, the system outputs are restricted to single dialogue acts by definition of the action space.  Each MDP action leads to a single dialogue act and the actions are mutually exclusive by definition.  In the multi-dimensional version, this restriction is still in place through the current combination rules, but we now have a more flexible mechanism in which such restrictions can be lifted.  For encoding logical conflicts between dialogue act candidates from different dimensions, for example answers in the task dimension and negative feedback \cite{Keizer:2007ve}, some combination rules can be retained.  For dealing with strategic and stylistic issues when evaluating dialogue act candidates, an additional MDP agent could be introduced and optimised jointly with the dimension-specific MDP agents.

To support the coordination process during training and thus make training more stable and efficient, each of the MDP models could be extended with information about the actions selected by the other MDPs.  The more dependent the dimensions turn out to be, the more explicit coordination might be required for learning, which in the most extreme case would lead to a model that is equivalent to the one-dimensional model.  However, the design of the used dialogue act taxonomy is such that a high level of independence between the MDPs can be expected, and therefore only modest explicit coordination might be needed.

\section{Conclusion and Future Work}\label{sec:concl}

We have argued for a \mdim\ approach to spoken dialogue system development, in order to enable more efficient cross-domain adaptation.  As a proof-of-concept, we have presented a first implementation of our multi-dimensional statistical dialogue manager and illustrated our approach with initial experiments in simulation, demonstrating the feasibility of training transferable conversational skills using multi-agent reinforcement learning and using these to speed up training in a new domain.

In future work, we will extend our dialogue manager and user simulator to support a wider range of dialogue act combinations, without the restrictions used in the initial experiments.  This will require further investigation into training settings for the multi-agent reinforcement learning framework, including dimension-specific reward functions and explicit coordination between the agents.  As we expand the action sets of the MDP agents, their state spaces will also need to be expanded, and value function approximation for policy optimisation will need to be upgraded from linear models to for example deep neural networks \cite{Mrksic_ea-2015,Zhao:2016us}.  We are also building an end-to-end system for the restaurant and smart home domains, in order to demonstrate our results on real data and across domains.

\paragraph*{Acknowledgments}
The MaDrIgAL project is funded by the EPSRC (EP/N017536/1).

\bibliographystyle{eacl2017}
\bibliography{keizer_rieser_semdial2017}

\end{document}